\documentclass[sigconf,nonacm]{acmart}

\renewcommand\footnotetextcopyrightpermission[1]{}
\usepackage{booktabs}
\usepackage{multirow}
\usepackage{graphicx}
\usepackage{subcaption}
\usepackage{cleveref}
\usepackage{tikz}
\usetikzlibrary{arrows.meta,positioning}

\graphicspath{{outputs/plots/}}
\newcommand{\rqone}{RQ1}
\newcommand{\rqtwo}{RQ2}

\newcommand{\paperplotgap}{\hspace{0.01\textwidth}}

\title{Complexity-Guided Component-wise Initialization for Language Model Pretraining}

\author{Konstantin Garbers}
\authornotemark[1]
\affiliation{%
  \institution{Peking University}
  \city{Beijing}
  \country{China}
}
\email{konstantin.garbers25@stu.pku.edu.cn}

\author{Nicholas Oh}
\authornote{Both authors contributed equally to this research.}
\affiliation{%
  \institution{Peking University}
  \city{Beijing}
  \country{China}
}
\email{2501213380@stu.pku.edu.cn}

\begin{document}

\begin{abstract}
Pretrained language models often exhibit structured weight spectra, suggesting that training may repeatedly produce similar layerwise and component-wise organization.
We ask whether these recurring spectral patterns can be reused as an initialization signal for GPT-2-style language-model pretraining.
First, we analyze eleven pretrained GPT-2-style checkpoints that vary in size, language, tokenizer, and training corpus, measuring Frobenius norm and effective-rank entropy across layers and Transformer subcomponents.
The checkpoints show shared depth trends, especially increasing scale and stronger spectral concentration in residual-writing matrices.
We then construct initialization schemes that imitate the component-wise magnitudes and spectral profiles of pretrained models, and compare them with several weight initialization methods.
These initializers visibly change the model's structural spectral patterns, but the evaluation results do not show a corresponding performance advantage.
Pretrained-weight reuse remains competitive, while coarse spectral matching alone is not a reliable optimization strategy.
Our results suggest that pretrained spectra are useful diagnostics of trained model structure, but that effective reuse likely requires preserving richer information than component-wise scale and singular-value shape.
\end{abstract}


\maketitle
\section{Introduction}

Large language models are typically trained from randomly initialized weights, often drawn from Gaussian or related distributions. This choice is robust and architecture-agnostic: it does not assume prior knowledge about the data, the model size, or the internal structure that training will eventually produce. However, it also ignores a growing body of evidence suggesting that trained language models are not arbitrary points in parameter space. Across models, layers, and components, pre-trained transformers often exhibit recurring structural regularities, including characteristic spectral properties of their weight matrices~\cite{wang-tu-2020-rethinking, yin2025outlierweighedlayerwisesparsity,martin2021implicit}.

These regularities raise a natural question: if trained models repeatedly develop similar patterns, can some of these patterns be used before training begins? In principle, an initialization that already reflects common structure found in pre-trained models could reduce the burden on optimization. Instead of learning all structural properties from scratch, the model would start from weights whose scale, rank structure, or spectral shape better resembles those of trained transformers. At the same time, it is unclear whether such coarse spectral information is actually useful for pre-training. Spectral similarity may capture meaningful structure, but it may also discard the specific directions, correlations, and feature-level organization that make pre-trained weights effective.

In this paper, we study this question for GPT-2-style language models. We analyze the weight matrices of several pre-trained models using spectral tools, including the Frobenius norm and effective rank entropy~\cite{martin2021implicit}. Our goal is first to determine whether consistent spectral patterns appear across models that differ in size, language, tokenizer, and training corpus. We then test whether these patterns can be transferred into the initialization of a new model and whether such initialization improves pre-training dynamics.

\paragraph{Research questions and answers.}
Specifically, we ask the following questions.

\begin{center}
\small
\centering
\begin{tabular}{@{}p{0.12\columnwidth}p{0.39\columnwidth}p{0.39\columnwidth}@{}}
\toprule
& \textbf{Research question} & \textbf{Answer} \\
\midrule
\textbf{\rqone} & What spectral patterns appear in pre-trained GPT-2-style language models? & Pre-trained GPT-2-style models show recurring layerwise and component-wise spectral trends despite differences in size, language, tokenizer, and corpus. \\
\textbf{\rqtwo} & How does initialization based on these spectral patterns affect training dynamics? & Reusing pre-trained weights remains competitive, but copying coarse spectral shape or scale alone is not a reliable optimization strategy. \\
\bottomrule
\end{tabular}
\end{center}

The remainder of this paper is organized as follows. We first analyze existing pre-trained LLMs in \cref{sec:spectral-patterns} to identify recurring spectral patterns in their weights. We then propose and evaluate a spectral-pattern-based initialization method in \cref{sec:weight-initialization}. Finally, we discuss related work in \cref{sec:related-works} and conclude in \cref{sec:conclusion}.

\section{Spectral Patterns in Pre-trained LLMs}
\label{sec:spectral-patterns}
Spectral analysis is the analysis of the eigenvalues or singular values of a matrix. In the context of neural networks, spectral analysis can provide insight into the geometric and functional properties of a weight matrix. Large singular values indicate that the matrix strongly amplifies certain input directions, which is often associated with greater complexity or sensitivity~\cite{pmlr-v9-glorot10a,staats2026small,martin2021implicit}. We define spectral patterns in 
\cref{sec:spectral-metrics} and analyze pretrained GPT-2-style models in \cref{sec:pretrained-spectral-patterns} to identify shared spectral patterns in pre-trained LLMs.

\subsection{GPT-2 Block Notation}
\label{sec:gpt2-block-notation}
GPT-2 is a decoder-only Transformer: tokens are embedded into a residual stream, then passed through a stack of identical causal self-attention and MLP blocks before the final language-model head.
Each block first applies layer normalization and multi-head causal self-attention, writes the attention output back to the residual stream, then applies a second layer normalization and a position-wise MLP with another residual write.
For block $\ell$, let $x_\ell$ denote the residual stream entering the block.
Using the pre-layer-norm GPT-2 convention, the two residual writes are
\[
  a_\ell = \operatorname{Attn}(\operatorname{LN}_1(x_\ell); W_{QKV}, W_O),
  \qquad
  \tilde{x}_\ell = x_\ell + a_\ell,
\]
\[
  m_\ell = W_{\mathrm{down}}\phi(W_{\mathrm{up}}\operatorname{LN}_2(\tilde{x}_\ell)),
  \qquad
  x_{\ell+1} = \tilde{x}_\ell + m_\ell,
\]
where $\phi$ is the MLP nonlinearity.
We focus on the four dense matrices that dominate the block computation.
The fused attention input matrix $W_{QKV}$ maps the residual stream into query, key, and value features; in GPT-2 implementations this is often stored as one combined projection and then split into $Q$, $K$, and $V$~\cite{radford2019language,wolf-etal-2020-transformers}.
The attention output matrix $W_O$ maps the concatenated head outputs back into the residual stream.
The MLP up-projection $W_{\mathrm{up}}$ expands the residual dimension into the hidden MLP dimension, while the MLP down-projection $W_{\mathrm{down}}$ maps the activated hidden representation back into the residual stream.
We summarize the component-weight and singular-value associations used in the component-wise interpretation in \cref{tab:block-component-associations}.
When we refer to a block-level aggregate, we combine these component matrices within a Transformer block; when we refer to a component plot, we track one of these matrices across depth.

\begin{table}[tbp]
  \centering
  \small
  \begin{tabular}{@{}p{0.22\columnwidth}p{0.70\columnwidth}@{}}
    \toprule
    Component & Association used in interpretation \\
    \midrule
    $W_Q, W_K$ & Larger singular values are associated with stronger feature directions for attention addressing~\cite{3618408.3620117,francoSingularVectorsAttention2026}. \\
    $W_V, W_O$ & Larger singular values are associated with stronger feature directions for transported content and stronger write strength to the residual stream~\cite{elhage2021mathematical,staats2026small,francoSingularVectorsAttention2026}. \\
    $W_{\mathrm{up}}$ & Component weights are usually associated with MLP key weights or feature-detector weights~\cite{gevaTransformerFeedForwardLayers2021,gevaTransformerFeedForwardLayers2022}. \\
    $W_{\mathrm{down}}$ & Component weights are usually associated with MLP value weights or stored-feature readout weights~\cite{gevaTransformerFeedForwardLayers2021,gevaTransformerFeedForwardLayers2022}. \\
    \bottomrule
  \end{tabular}
  \caption{Interpretive associations for component weights and singular values in Transformer block components.\protect\footnotemark}
  \Description{A table listing W Q and W K, W V and W O, W up, and W down with the component-weight or singular-value association used for each component.}
  \label{tab:block-component-associations}
\end{table}
\footnotetext{In the later experiments we consider $W_{QKV}$ jointly and $W_O$ separately. This grouping is a limitation, because the functionality of $W_V$ is more closely related to $W_O$ than to $W_Q$ and $W_K$.}

\begin{figure}[tbp]
  \centering
  \resizebox{0.98\columnwidth}{!}{%
  \begin{tikzpicture}[
    font=\scriptsize,
    node distance=4.5mm and 5.5mm,
    box/.style={draw, rounded corners=1pt, minimum height=6mm, align=center, inner xsep=3pt},
    add/.style={circle, draw, inner sep=0.5pt, minimum size=4.5mm},
    line/.style={-{Latex[length=1.5mm]}, thick}
  ]
    \node (x) {$x_\ell$};
    \node[box, right=of x] (ln1) {LN$_1$};
    \node[box, right=of ln1] (attn) {$W_{QKV}$\\Attn\\$W_O$};
    \node[add, right=of attn] (add1) {$+$};
    \node[box, right=of add1] (ln2) {LN$_2$};
    \node[box, right=of ln2] (mlp) {$W_{\mathrm{up}}$\\$\phi$\\$W_{\mathrm{down}}$};
    \node[add, right=of mlp] (add2) {$+$};
    \node[right=of add2] (out) {$x_{\ell+1}$};

    \draw[line] (x) -- (ln1);
    \draw[line] (ln1) -- (attn);
    \draw[line] (attn) -- (add1);
    \draw[line] (add1) -- node[above] {$\tilde{x}_\ell$} (ln2);
    \draw[line] (ln2) -- (mlp);
    \draw[line] (mlp) -- (add2);
    \draw[line] (add2) -- (out);
    \draw[line] (x) to[out=65,in=115] (add1);
    \draw[line] (add1) to[out=65,in=115] (add2);
  \end{tikzpicture}
  }
  \caption{Pre-layer-norm GPT-2 block notation. Attention and MLP outputs are added back into the residual stream.}
  \Description{A compact GPT-2 block diagram showing x l flowing through layer normalization, attention, residual addition, layer normalization, MLP, residual addition, and x l plus one.}
  \label{fig:gpt2-block-notation}
\end{figure}
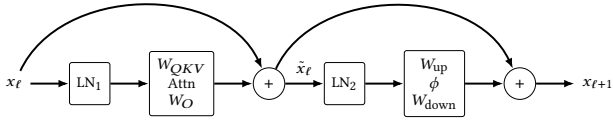

\subsection{Spectral Analysis Metrics}
\label{sec:spectral-metrics}
We use static spectral metrics because they are inexpensive to compute directly from saved weight matrices and do not require additional data passes, interventions, or fine-tuning runs.
They are also established diagnostics for trained neural networks and Transformer weights: prior work uses spectral summaries to study implicit regularization, layer and component structure, initialization scale, and complexity control~\cite{martin2021implicit,wang-tu-2020-rethinking,yin2025outlierweighedlayerwisesparsity,staats2026small,3737916.3738367,11304601}.
Our chosen metrics separate complementary properties of each matrix: Frobenius norm captures total scale, while effective-rank entropy describes how concentrated or distributed that scale is across singular directions.

\paragraph{Frobenius Norm}
The Frobenius norm is
\[
  \lVert W\rVert_F = \sqrt{\sum_{i,j} W_{ij}^2}
  = \sqrt{\sum_k \sigma_k(W)^2}.
\]
It measures total weight energy and is useful for separating changes in scale from changes in spectral shape.

\paragraph{Effective Rank Entropy}
Frobenius norm alone does not capture the distribution of singular values. Let $\sigma_i$ be the singular values of $W$, and let $p_i=\sigma_i/\sum_j \sigma_j$.
We use the entropy effective rank
\[
  \operatorname{erank}(W) = \exp\left(-\sum_i p_i \log p_i\right).
\]
This measures how many singular directions carry substantial mass~\cite{martin2021implicit}.

\subsection{Pre-trained LLMs Share Similar Spectral Patterns} 
\label{sec:pretrained-spectral-patterns}
We analyze eleven Hugging Face checkpoints: English GPT-2 small and medium models~\cite{radford2019language}, Russian GPT-2 small and large models \cite{zmitrovich-etal-2024-family}, and GPT-2-style Vietnamese, Chinese, Portuguese, Japanese, Turkish, poem-generation, and story-generation models \cite{NlpHUSTGpt2vietnameseHugging2025,MymusiseGpt2mediumchineseHugging,pierre2020gpt2smallportuguese,rinna-japanese-gpt2-medium,sawada-etal-2024-release,kesgin2024introducing,zhao2023tencentpretrain,PranavpsvGpt2genrestorygeneratorHugging}.
The selection keeps the architecture family fixed while varying size, language, tokenizer, and training corpus.
We analyze block-level aggregates and four subcomponents: the fused $W_{QKV}$ attention input projection, the $W_O$ attention output projection, the $W_{\mathrm{up}}$ MLP up projection, and the $W_{\mathrm{down}}$ MLP down projection.
We treat $W_{QKV}$ jointly because it determines the query, key, and value feature spaces used to compute and transport attention information.
We analyze $W_O$ separately because it maps attention outputs back into the residual stream, so its spectrum is more directly tied to write-back strength and directionality than to attention-score formation.

\paragraph{General Observations}
Across pretrained checkpoints, layer-level Frobenius norm and effective-rank entropy follow similar trends after per-model normalization.
The shared shape is clearer in z-score plots than metric-value plots, indicating that the common signal is mostly structural rather than a consequence of identical absolute scale.
At the component level, effective-rank entropy drops across multiple components in the final few layers.
The decrease is most pronounced for the residual-writing or value-side matrices $W_O$ and $W_{\mathrm{down}}$, but a similar late-depth decrease also appears for $W_{QKV}$ and $W_{\mathrm{up}}$.
This means that the represented spectrum becomes more concentrated near the top of the network: fewer singular directions carry a large share of the spectral mass.
At the same time, the Frobenius norms of $W_O$ and $W_{\mathrm{down}}$ rise almost linearly with depth, implying that the total weight mass of these matrices increases while their effective spectra become narrower.
The $W_{QKV}$ and $W_{\mathrm{up}}$ matrices follow related but weaker depth shapes; common and individual trends are less pronounced there than for $W_O$ and $W_{\mathrm{down}}$.

\begin{figure*}[tbp]
  \centering
  \begin{subfigure}[t]{0.23\textwidth}
  \centering
  \includegraphics[width=\linewidth]{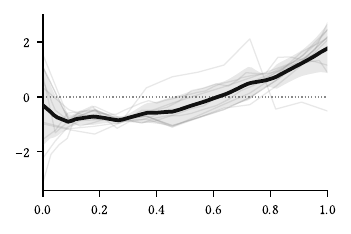}
  \caption{Layer Frobenius norm.}
  \Description{Layer Frobenius norm.}
  \label{fig:pretrained-frobenius-norm}
\end{subfigure}\paperplotgap
  \begin{subfigure}[t]{0.23\textwidth}
  \centering
  \includegraphics[width=\linewidth]{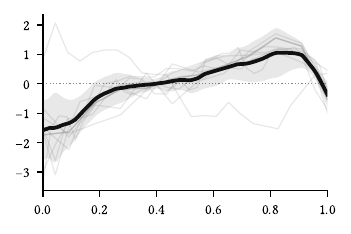}
  \caption{Layer effective entropy.}
  \Description{Layer effective entropy.}
  \label{fig:pretrained-effective-rank}
\end{subfigure}
  \caption{Pretrained layerwise Frobenius-norm and effective-entropy trends.}
  \Description{Pretrained layerwise Frobenius-norm and effective-entropy trends.}
\end{figure*}

\paragraph{Interpretation}
We interpret the component-level trends through the idea of neural memories~\cite{sukhbaatar2015endtoend}.
In this view, a two-matrix computation can be read as a key-value memory: one matrix or matrix product determines which features are addressed, while the other determines what information is returned.
This interpretation has been applied to Transformer feed-forward layers because their up- and down-projections form such a two-matrix computation~\cite{gevaTransformerFeedForwardLayers2021}.
We can apply the same key-value interpretation to the attention weight matrices.

Under this view, the $W_{QK}$ weights function similarly to keys because they determine the attention logits and therefore the attention weights that select or mix token positions~\cite{elhage2021mathematical}.
The $W_{VO}$ weights function as the value pathway because they transport the selected content back into the residual stream.
This gives a natural interpretation of the weaker Frobenius-norm trend for $W_{QKV}$: changes in attention-logit magnitude are normalized when logits are converted into attention weights explaining the almost constant Frobenius norm, while the late-layer Frobenius-norm growth and effective-entropy drop in $W_O$ suggest increasingly strong but more spectrally concentrated residual write-back directions.

The MLP pattern can then be read through the original up-and-down key-value interpretation: $W_{\mathrm{up}}$ detects or keys features in the hidden MLP space, while $W_{\mathrm{down}}$ writes value-like information back to the residual stream~\cite{gevaTransformerFeedForwardLayers2021,gevaTransformerFeedForwardLayers2022}.
Because MLP layers are often associated with factual and feature storage~\cite{gevaTransformerFeedForwardLayers2021,gevaTransformerFeedForwardLayers2022,mengLocatingEditingFactual2022}, the stronger common trends in $W_{\mathrm{down}}$ may reflect later-layer specialization in the readout or residual-writing side of the MLP.
Taken together, the strongest shared spectral trends appear in the matrices most directly responsible for writing information into the residual stream, although this should be understood as a qualitative interpretation of the plots rather than a conclusion from a separate quantitative analysis.

\begin{figure*}[tbp]
  \centering
  \begin{subfigure}[t]{0.23\textwidth}
  \centering
  \includegraphics[width=\linewidth]{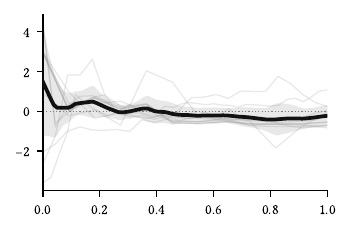}
  \caption{$W_{QKV}$ Frobenius norm.}
  \Description{$W_{QKV}$ Frobenius norm.}
  \label{fig:pretrained-qkv-frobenius-norm}
\end{subfigure}\paperplotgap
  \begin{subfigure}[t]{0.23\textwidth}
  \centering
  \includegraphics[width=\linewidth]{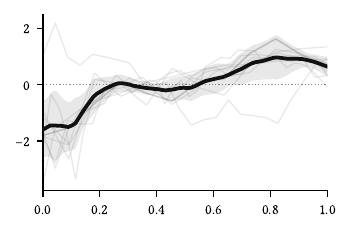}
  \caption{$W_{QKV}$ effective entropy.}
  \Description{$W_{QKV}$ effective entropy.}
  \label{fig:pretrained-qkv-effective-rank}
\end{subfigure}\paperplotgap
  \begin{subfigure}[t]{0.23\textwidth}
  \centering
  \includegraphics[width=\linewidth]{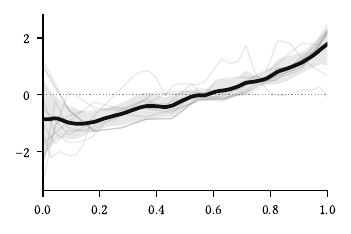}
  \caption{$W_O$ Frobenius norm.}
  \Description{$W_O$ Frobenius norm.}
  \label{fig:pretrained-wo-frobenius-norm}
\end{subfigure}\paperplotgap
  \begin{subfigure}[t]{0.23\textwidth}
  \centering
  \includegraphics[width=\linewidth]{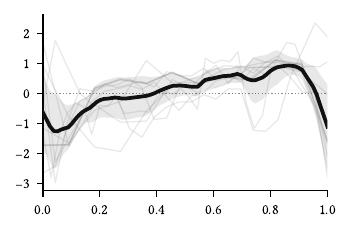}
  \caption{$W_O$ effective entropy.}
  \Description{$W_O$ effective entropy.}
  \label{fig:pretrained-wo-effective-rank}
\end{subfigure}
  \par\medskip
  \begin{subfigure}[t]{0.23\textwidth}
  \centering
  \includegraphics[width=\linewidth]{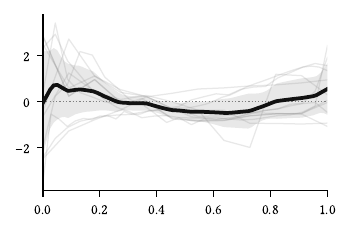}
  \caption{$W_{\mathrm{up}}$ Frobenius norm.}
  \Description{$W_{\mathrm{up}}$ Frobenius norm.}
  \label{fig:pretrained-mlp-up-frobenius-norm}
\end{subfigure}\paperplotgap
  \begin{subfigure}[t]{0.23\textwidth}
  \centering
  \includegraphics[width=\linewidth]{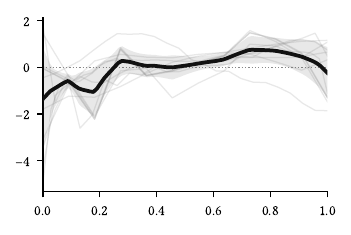}
  \caption{$W_{\mathrm{up}}$ effective entropy.}
  \Description{$W_{\mathrm{up}}$ effective entropy.}
  \label{fig:pretrained-mlp-up-effective-rank}
\end{subfigure}\paperplotgap
  \begin{subfigure}[t]{0.23\textwidth}
  \centering
  \includegraphics[width=\linewidth]{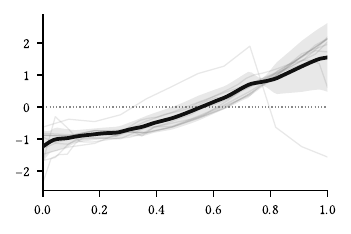}
  \caption{$W_{\mathrm{down}}$ Frobenius norm.}
  \Description{$W_{\mathrm{down}}$ Frobenius norm.}
  \label{fig:pretrained-mlp-down-frobenius-norm}
\end{subfigure}\paperplotgap
  \begin{subfigure}[t]{0.23\textwidth}
  \centering
  \includegraphics[width=\linewidth]{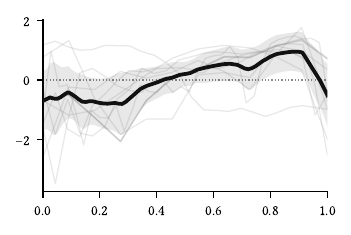}
  \caption{$W_{\mathrm{down}}$ effective entropy.}
  \Description{$W_{\mathrm{down}}$ effective entropy.}
  \label{fig:pretrained-mlp-down-effective-rank}
\end{subfigure}
  \caption{Pretrained component-wise Frobenius-norm and effective-entropy trends for attention and MLP projections.}
  \Description{A two-row grid of component-wise pretrained spectral diagnostics. The first row shows W QKV and W O Frobenius norm and effective entropy. The second row shows W up and W down Frobenius norm and effective entropy.}
  \label{fig:pretrained-component-trends}
\end{figure*}

\section{Spectral-pattern-based Weight Initialization}
\label{sec:weight-initialization}
We now test whether the spectral patterns identified in pretrained GPT-2-style models can be turned into useful initialization rules for training a new model.

\subsection{Proposed Solution}
We turn the pretrained depth trends from \cref{sec:spectral-patterns} into component-wise initialization rules.
The proposed initializers are deliberately coarse: they approximate broad scale and spectral concentration profiles rather than fitting every checkpoint-specific curve.
This design tests whether pretrained-like scale and singular-value shape are useful initialization signals when the model architecture is fixed.
The proposed initializers and comparison baselines are summarized in \cref{tab:init-methods}.

\begin{table*}[tbp]
  \centering
  \caption{Initialization methods compared in the training experiments.}
  \label{tab:init-methods}
  \Description{Table describing proposed spectral-pattern initialization methods and comparison baselines.}
  \begin{tabular}{llp{0.70\linewidth}}
    \toprule
    Type & Method & Construction \\
    \midrule
    \multirow{4}{*}{Baseline} & Standard & Samples linear and embedding weights from a zero-mean Gaussian with standard deviation $0.02$ and applies GPT-2 residual-branch scaling~\cite{radford2019language}. \\
    & High std & Keeps the standard GPT-2 recipe but increases the initialization standard deviation to $0.08$. \\
    & No residual scaling & Uses the standard Gaussian scale but removes GPT-2 residual-branch scaling. \\
    & Pretrained reuse & Copies non-token-embedding weights from a Chinese GPT-2-medium checkpoint and reinitializes token embeddings for the GPT-2 tokenizer~\cite{MymusiseGpt2mediumchineseHugging}. \\
    \midrule
    \multirow{3}{*}{Proposed} & Magnitude & Replaces the single global scale with per-layer, per-component target standard deviations: $W_{QKV}$ decreases with depth, $W_{\mathrm{up}}$ decreases mildly, and residual-writing projections increase with depth. \\
    & Mag.+spectrum & Uses the same target magnitudes, then tries to match the stronger spectral concentration seen in pretrained models by smoothly decaying the singular values. Concretely, it multiplies singular values by $\exp(-\lambda(z)i/n)$, where $\lambda(z)=5(1-z)$ and $z$ is relative depth, before rescaling each matrix to $\lVert W\rVert_F=\sqrt{d_{\mathrm{in}}d_{\mathrm{out}}}\sigma(z)$. \\
    & Realistic scale & Applies the magnitude-spectrum construction and then rescales each subcomponent with cohort-derived multipliers so the predicted Frobenius RMS matches the trained cohort scale per subcomponent. \\
    \bottomrule
  \end{tabular}
\end{table*}

\begin{figure*}[tbp]
  \centering
  \includegraphics{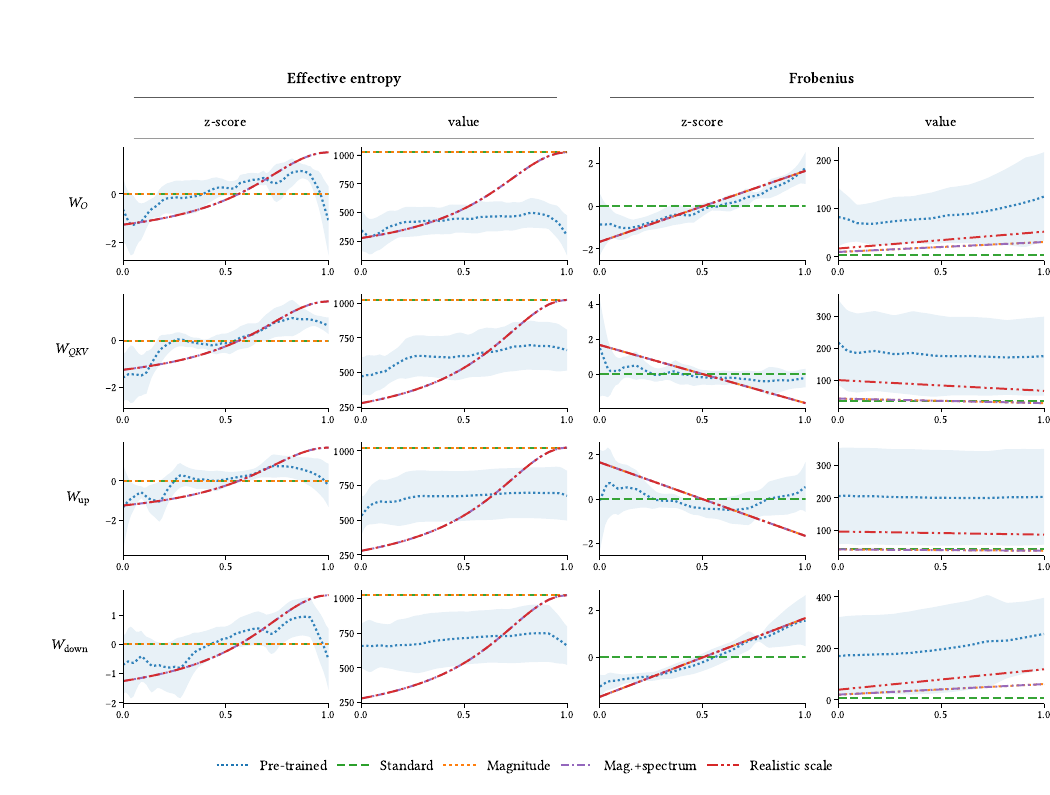}
  \caption{Initialization diagnostic grid. Rows correspond to $W_O$, $W_{QKV}$, $W_{\mathrm{up}}$, and $W_{\mathrm{down}}$. Columns group panels by metric family, effective entropy or Frobenius norm, and by value type, z-score or metric value. In each panel, relative depth is shown on the horizontal axis and the corresponding diagnostic value is shown on the vertical axis. Line styles distinguish the initialization strategies, while the blue dotted curve and shaded band show the pre-trained cohort mean and $\pm1$ standard deviation.}
  \Description{Initialization diagnostic grid comparing pre-trained cohort trends with Standard, Magnitude, Mag.+spectrum, and Realistic scale initialization curves.}
  \label{fig:init-grid-entropy-frobenius}
\end{figure*}

The spectrum-shaping rule separates singular-value shape from total magnitude through the final Frobenius rescaling.
Because it uses the normalized singular-value index $i/n$, the same rule can be applied across component matrices with different shapes.
The resulting initialized profiles reproduce some coarse pretrained tendencies in Frobenius norm and effective-rank entropy.
\Cref{fig:init-grid-entropy-frobenius} shows these initialization profiles against the pretrained cohort summary.

\subsection{Results}
\paragraph{Experiment Setup}
All full runs use the same 24-layer GPT-2-medium-sized architecture ($d_{\mathrm{model}}=1024$, 16 heads, context length 1024) with the GPT-2 tokenizer and SlimPajama-6B~\cite{radford2019language,cerebras2023slimpajama}.
We train for 90k steps, approximately one epoch over 6B tokens, on 8 H20 GPUs with seed 1, per-GPU batch size 2, gradient accumulation 4, learning rate $3\cdot10^{-4}$, weight decay 0.1, 10M validation tokens, evaluation every 1000 steps, and checkpoints every 10k steps.
We evaluate validation perplexity and held-out perplexity on FineWeb-Edu, OpenWebText, and WikiText-103~\cite{NEURIPS2024_370df50c,lozhkov2024fineweb-edu,Gokaslan2019OpenWeb,merity2016pointer}.
We also evaluate BLiMP syntactic minimal-pair accuracy~\cite{warstadt2020blimp} and multiple-choice accuracy on ARC-Challenge, ARC-Easy, and WinoGrande \cite{clark2018thinksolvedquestionanswering,sakaguchi2019winograndeadversarialwinogradschema}.
We compare the proposed initializers with Standard, High std, No residual scaling, and Pretrained reuse baselines.

\paragraph{Empirical Evaluation}
The proposed initializers do not show a uniform benefit over the baselines.
Magnitude hurts BLiMP accuracy, while Mag.+\allowbreak spectrum has the weakest validation and held-out perplexities.
Realistic scale narrows this gap, suggesting that absolute scale is important, but it still does not dominate the Standard baseline.
Pretrained reuse is competitive on perplexity, BLiMP, WikiText-103, and ARC-Easy despite the tokenizer/language mismatch, but it is weak on ARC-Challenge.
Overall, pretrained spectra are descriptive of trained models, but not sufficient by themselves as an optimization strategy.
\Cref{tab:evaluation-summary} summarizes the evaluation results.

\begin{table*}[tbp]
  \centering
  \caption{Evaluation summary grouped by criterion. Arrows indicate whether lower or higher values are better. PPL denotes perplexity, Acc.\ \% denotes BLiMP syntactic minimal-pair accuracy, and MC \% denotes multiple-choice accuracy. Bold marks the best result in a column, and red marks clear negative outliers among the full 90k-step runs. Proposed initializers are separated from baselines by a horizontal rule.}
  \label{tab:evaluation-summary}
  \Description{Evaluation summary table with arrows for metric direction. PPL is perplexity, where lower is better; Acc. percent is BLiMP accuracy, where higher is better; MC percent is multiple-choice accuracy, where higher is better.}
  \begin{tabular}{lrrrrrrrr}
    \toprule
    & \multicolumn{4}{c}{PPL $\downarrow$} & \multicolumn{1}{c}{Acc.\ \% $\uparrow$} & \multicolumn{3}{c}{MC \% $\uparrow$} \\
    \cmidrule(lr){2-5}\cmidrule(lr){6-6}\cmidrule(lr){7-9}
    Init & Val & FineWeb & OWT & WikiText & BLiMP & ARC-C & ARC-E & WinoG \\
    \midrule
    Standard & \textbf{21.68} & \textbf{29.9} & \textbf{30.3} & 51.9 & 80.0 & 24.2 & 37.4 & 50.4 \\
    High std & 22.20 & 30.7 & 31.3 & 51.9 & \textbf{80.9} & 23.1 & 38.0 & 50.3 \\
    No residual scaling & 21.95 & 30.3 & 31.0 & 53.2 & 80.8 & 23.5 & 38.0 & \textbf{51.1} \\
    Pretrained reuse & \textbf{21.68} & \textbf{29.9} & 30.7 & \textbf{49.4} & \textbf{80.9} & \textcolor{red}{21.8} & \textbf{38.6} & 50.3 \\
    \midrule
    Magnitude & 22.06 & 30.5 & 31.2 & 51.3 & \textcolor{red}{76.2} & 23.3 & 38.1 & 50.6 \\
    Mag.+spectrum & \textcolor{red}{22.24} & \textcolor{red}{30.8} & \textcolor{red}{31.5} & \textcolor{red}{53.8} & 80.6 & 22.6 & 38.4 & 50.4 \\
    Realistic scale & 21.93 & 30.3 & 31.2 & 52.1 & 80.0 & 22.6 & 37.6 & 50.9 \\
    \bottomrule
  \end{tabular}
\end{table*}

\begin{figure*}[tbp]
  \centering
  \includegraphics{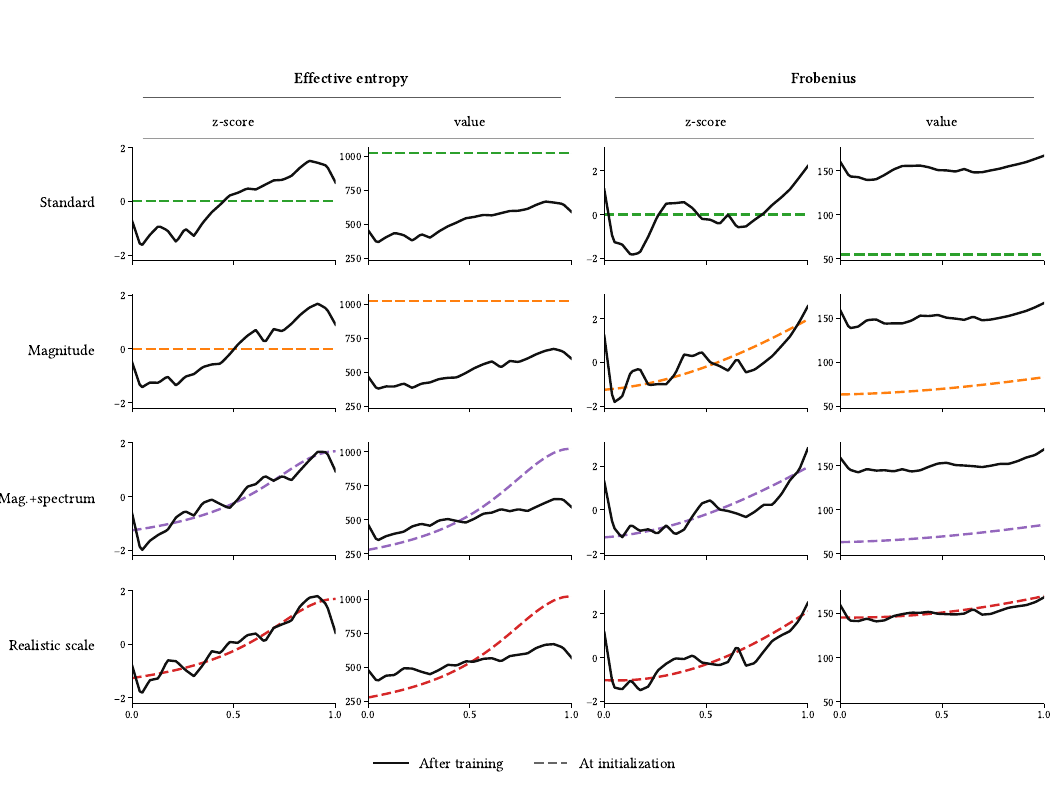}
  \caption{Strategy-matched initialization and final-checkpoint diagnostics. Rows correspond to the initialization strategy used for training. Columns follow the same structure as the main initialization grid: effective entropy and Frobenius norm, each shown as z-score and metric value. Dashed colored curves show the closed-form initialization prediction, while solid black curves show the corresponding trained checkpoint.}
  \Description{A four-row by four-column grid comparing each initialization strategy to the trained checkpoint produced by that same strategy. Columns show effective entropy z-score and value, followed by Frobenius norm z-score and value.}
  \label{fig:init-vs-trained-spectral}
\end{figure*}

\paragraph{Spectral Analysis}
The proposed initializers visibly change the structural spectral patterns of the model.
The magnitude-based initializers impose clear Frobenius-norm depth profiles at initialization, with decreasing scale for $W_{QKV}$ and $W_{\mathrm{up}}$ and increasing scale for the residual-writing matrices $W_O$ and $W_{\mathrm{down}}$.
Mag.+\allowbreak spectrum additionally changes effective-entropy structure by making the initialized spectra more concentrated.
After training, these imposed profiles are partly overwritten but not fully erased: the final checkpoints move toward trained-model-like curves while retaining strategy-dependent differences, especially in component-wise scale and spectral concentration.
\Cref{fig:init-vs-trained-spectral} compares each strategy's initialization profile with the corresponding trained checkpoint.

\section{Discussion}
We discuss the main limitations of our spectral analysis and initialization experiments, and outline directions for testing richer forms of pretrained structure reuse.

\subsection{Threats to Validity}
\paragraph{External Validity}
Our analysis is limited to GPT-2-style decoder-only Transformers.
This constraint keeps the architecture family fixed and makes component-wise comparisons cleaner, but it also limits how far the conclusions can be transferred to other model families, attention variants, normalization schemes, or larger contemporary LLMs.
Within this family, we partially mitigate the limitation by sampling checkpoints with different sizes, languages, tokenizers, and training corpora.
\paragraph{Internal Validity}
The training experiments use one main random seed and one primary pretraining dataset.
As a result, small differences between initialization strategies may reflect seed-specific optimization noise or dataset-specific effects rather than stable differences in the methods.
The strongest negative results are therefore more reliable than fine-grained rankings among runs with similar perplexity or accuracy.
\paragraph{Construct Validity}
The proposed initializers approximate broad component-wise spectral trends rather than reproducing every detail of the pretrained spectra.
This means they can miss local features such as sharp dips, end-of-depth spikes, or short-range changes that may be functionally important.
Our results therefore test whether coarse spectral shape and scale are useful initialization signals, not whether an exact spectral replica of a pretrained model would improve training.
\subsection{Future Work}
Future work should separate the descriptive role of spectral complexity from its causal effect on downstream performance.
In this paper, complexity is partly enforced through initialization, so it remains unclear whether naturally arising changes in effective rank or Frobenius norm predict better generalization when they are not explicitly imposed.
A broader study could measure how these diagnostics correlate with performance across model sizes, datasets, checkpoints, and training stages.

Another direction is to study which pretrained patterns are shared across language models and which are specific to a model, language, or corpus. Mechanistic-interpretability work often builds on the linear representation hypothesis, under which high-level features are encoded as directions in model representation spaces; recent work further asks whether such directions can be aligned or transferred across models~\cite{pmlr-v235-huh24a,3780338.3782228,huang-etal-2025-cross}. Our results suggest an analogous question at the weight level: whether repeated component-wise spectral patterns are merely consequences of architecture and optimization, or whether they encode reusable structure for new training runs.

The competitive performance of pretrained-weight reuse, despite a tokenizer and language mismatch, suggests that some form of pattern reuse is possible.
However, our spectral initializers show that coarse spectrum matching is not enough to capture the benefit.
Future initialization methods should test richer reuse signals, such as preserving subspace geometry, component-specific singular vectors, or recurring layer-specific weight patterns, and should evaluate whether any early training advantage persists at longer training horizons.

\section{Related Work}
\label{sec:related-works}
This work connects to four lines of prior work.
\paragraph{Weight Initialization}
Classical initialization methods control activation and gradient scale at the start of training~\cite{pmlr-v9-glorot10a,he2015delving}. In Transformer language models, initialization also interacts with residual-depth scaling: GPT-2 scales residual-layer weights at initialization to compensate for accumulation across many residual branches~\cite{radford2019language}. More recent work studies how initialization scale and spectral control affect training stability and whether learned solutions favor reasoning-like or memorization-like behavior~\cite{3618408.3620117,3737916.3738367,11304601}. Our experiments keep the architecture fixed and ask whether pretrained component-wise spectral profiles can provide a useful initialization signal beyond global variance choices in NLP tasks.
\paragraph{Spectral Structure in Trained Models}
Spectral analysis has been used to characterize implicit regularization and complexity in trained weights~\cite{martin2021implicit}. Related analyses of Transformer components show that layers and subcomponents can differ systematically in importance or retained information~\cite{wang-tu-2020-rethinking,yin2025outlierweighedlayerwisesparsity,staats2026small}. Our first-stage analysis follows this diagnostic view, but focuses on whether GPT-2-style checkpoints share recurring component-wise depth profiles across model size, language, tokenizer, and corpus.
\paragraph{Mechanistic Interpretability}
Mechanistic-interpretability work motivates interpreting Transformer components as structured computations: attention can be decomposed into query-key and output-value circuits~\cite{elhage2021mathematical}, MLP blocks can behave like key-value memories or vocabulary-space updates~\cite{gevaTransformerFeedForwardLayers2021,gevaTransformerFeedForwardLayers2022,mengLocatingEditingFactual2022}, and feature directions may be recoverable from singular-vector structure~\cite{francoSingularVectorsAttention2026,staats2026small}. This motivates our component-wise treatment of $W_{QKV}$, $W_O$, $W_{\mathrm{up}}$, and $W_{\mathrm{down}}$, although our measurements remain spectral rather than causal or circuit-level.
\paragraph{Complexity Control and Generalization}
Spectral and low-rank structure is also widely used for model compression and resource allocation in Transformer language models~\cite{hua-etal-2023-dynamic,wang-etal-2025-svd-llm}. This line of work treats singular values as a practical handle on model size and retained information, whereas our experiments ask whether pretrained spectral structure can be reused as an initialization signal. Work on cross-model representation alignment suggests that internal feature spaces can sometimes be compared or transferred across models~\cite{pmlr-v235-huh24a,3780338.3782228,huang-etal-2025-cross}; our setting asks a narrower weight-level version of this question, where only coarse spectral summaries are transferred.

\section{Conclusion}
\label{sec:conclusion}
We studied whether spectral patterns found in pretrained GPT-2-style models can be transferred into the initialization of a new language model.
Across eleven pretrained checkpoints, we found recurring layerwise and component-wise trends despite differences in model size, language, tokenizer, and corpus.
The most consistent patterns appear in residual-writing components: $W_O$ and $W_{\mathrm{down}}$ tend to grow in Frobenius norm with depth while their effective spectra become more concentrated.

Our initialization experiments show that these patterns can be imposed structurally.
Initializers that replace the single global weight scale with layer- and component-specific magnitudes, and that additionally reshape singular values to mimic the stronger spectral concentration of pretrained models, visibly alter the model's Frobenius-norm and effective-entropy profiles; some of these differences remain after training.
However, these structural changes do not translate into a consistent evaluation gain.
The magnitude-only weight initialization intervention hurts BLiMP accuracy, adding singular-value reshaping performs worst on several perplexity measures, and matching pretrained absolute scale more closely narrows but does not remove the gap to standard initialization.
By contrast, direct pretrained-weight reuse remains competitive despite a tokenizer and language mismatch, indicating that useful transferable structure may exist but is not captured by coarse spectral summaries alone.

The main implication is therefore negative but informative: pretrained spectra describe real regularities in trained Transformer weights, yet copying component-wise scale and singular-value shape is insufficient as a standalone pretraining initialization method.
Future work should test richer ways of reusing pretrained structure, such as singular-vector information and component-specific patterns, and should study whether spectral metrics only describe trained models or can directly improve optimization and generalization.

\bibliographystyle{ACM-Reference-Format}
\bibliography{references}

\end{document}